\pdfoutput=1
\documentstyle[twocolumn,graphicx]{article}

\begin{document}

\date{}

\title{\bf Accelerating and Evaluation of Syntactic Parsing in Natural Language Question Answering Systems
\thanks{In Proceedings of International Conference on Artificial Intelligence (ICAI'07), pp. 595-601, Las Vegas, Nevada, USA, June 2007.}}

\author{
{\bf Zhe Chen}\\
BASICS Lab\\
Department of Computer Science\\
Shanghai Jiao Tong University\\
Shanghai 200240, China\\
\and
{\bf Dunwei Wen}\\
\\
School of Computing and Information Systems\\
Athabasca University\\
Athabasca AB T9S 3A3 Canada\\
}

\maketitle                        
\thispagestyle{empty}


\noindent {\bf Abstract {\small\em With the development of Natural
Language Processing (NLP), more and more systems want to adopt NLP
in User Interface Module to process user input, in order to
communicate with user in a natural way. However, this raises a speed
problem. That is, if NLP module can not process sentences in durable
time delay, users will never use the system. As a result, systems
which are strict with processing time, such as dialogue systems, web
search systems, automatic customer service systems, especially
real-time systems, have to abandon NLP module in order to get a
faster system response. This paper aims to solve the speed problem.
In this paper, at first, the construction of a syntactic parser
which is based on corpus machine learning and statistics model is
introduced, and then a speed problem analysis is performed on the
parser and its algorithms. Based on the analysis, two accelerating
methods, Compressed POS Set and Syntactic Patterns Pruning, are
proposed, which can effectively improve the time efficiency of
parsing in NLP module. To evaluate different parameters in the
accelerating algorithms, two new factors, PT and RT, are introduced
and explained in detail. Experiments are also completed to prove and
test these methods, which will surely contribute to the application
of NLP. } }

\vspace{0.5cm}



\section{Introduction}
Natural Language Processing (NLP) is one of the most important
fields in Artificial Intelligence researches, and it is applied more
and more in application systems. For example, NLP could be used in
Question Answering (QA) systems to understand users' natural
language inputs, and communicate with users in a natural way, such
as LUNAR \cite{Woods1977} and some service systems \cite{Cheng2002}.
These applications have greatly improved the way users interact with
computer systems and overcome the disadvantages of traditional QA
systems which use pattern matching algorithms, for example ALICE
\cite{Wallace2006}.

However, with the development of NLP technology, a big problem has
emerged. Most researchers spend a lot of time thinking of how to
improve the precision of Part-of-Speech (POS) taggers and syntactic
parsers, but there are few researches on how to save CPU time in
tagging and parsing without precision decrease. Actually nowadays,
NLP is applied more and more in real-time QA systems, such as
dialogue, web search, cell phone and PDA etc. \cite{Chen2005}. As a
result, the processing time problem becomes more and more important
for NLP applications, because users need the responses to their
requests in an acceptable length of time. In fact, the speed problem
is the very reason why most QA systems choose pattern matching
algorithm but not NLP methods.

Then, how to accelerate the parsing speed of syntactic parser? A NLP
system always includes several parts, such as a stemmer module, a
word tagging module, and a syntactic parsing module etc. Many
algorithms have been proposed for these modules. As we know, the
syntactic parsing takes most of the processing time. So, improving
syntactic parsing is one of the most important methods, and the
optimization of other modules is also necessary.

Syntactic patterns are needed in syntactic parsing module. But it is
possible for humans to construct a syntactic pattern. Firstly, it is
hard to define a large amount of syntactic patterns. Secondly, it is
impossible to decide the probability of each pattern's appearance.
So corpus machine learning algorithm would be the best way to
generate syntactic pattern dictionary. In this paper, we use Penn
Corpus \cite{Penn} developed by Penn University in our research.
Penn Treebank project \cite{Marcus1993} produces skeletal parses
based on an initial POS tagging showing rough syntactic and semantic
information on about 2.5 million English words.

In the rest of this paper, we will introduce the problems which
exist in constructing syntactic parser first. Then an improved
corpus learning algorithm will be proposed to improve the time
efficiency of parsing. To evaluate the time efficiency of parsing,
two new evaluation factors will be invented. Also experiments will
be done to prove these algorithms. At the end of this paper,
conclusions and future work are summarized.

\section{Syntactic Parser Construction}
In this section, we will introduce how to construct a parser which
can learn from corpus, discuss how to parse sentences, and analyze
the reason why the speed problem exists.

\subsection{Corpus Machine Learning}
Syntactic pattern learning generates syntactic pattern dictionary
through machine learning from tagged and parsed corpus. In the
learning process, all the appeared patterns should be extracted from
corpus, and also their appearance counts and probabilities should be
recorded for the further processing.

In this paper, $N$ stands for nonterminals, like
``S'',``NP'',``VP'', and $N^j$ means $j-th$ nonterminal in the
nonterminal set.

For each syntactic pattern $N^j \rightarrow \zeta$, where $N^j$ and
$\zeta$ are both Part-Of-Speech (POS), $C(N^j \rightarrow \zeta)$ is
used to record appearance count of the pattern, and $P(N^j
\rightarrow \zeta)$ represents its appearance probability. The
following Formula \ref{Equ:ML} represents the relationship between
the two variables:

\begin{equation}\label{Equ:ML}
P(N^j \rightarrow \zeta) = \frac{C(N^j \rightarrow
\zeta)}{\sum_\gamma C(N^j \rightarrow \gamma)}
\end{equation}

In this equation, $\gamma \in R$ and $R$ is the full POS Pattern
Set. So $\sum_\gamma C(N^j \rightarrow \gamma)$ stands for the total
amount of all the possible patterns which have the same left $N^j$.

\subsection{Parsing and Speed Problem}
Syntactic Parsing is defined to generate a syntactic tree form a
given sentence. For example, we can use chart parsing algorithm to
parse sentences, for more details, please see \cite{Russell2003},
\cite{Earley1970} and \cite{Moore2000}.

In the function ``predictor" of chart parsing, patterns with the
required left side $N^j$, like ``$N^j \rightarrow \gamma$", are all
added into chart to predict next matching pattern. However, if the
system has a large syntactic pattern dictionary, a lot of patterns
will be added, including both important patterns and unimportant
patterns which have low probability to appear. Actually in most
cases, the unimportant patterns will contribute nothing to the
parsing tree. In other words, it is most likely that the unimportant
patterns are not part of the syntactic tree, but they spend most
processing time in parsing. As a result, the system will spend much
time in processing meaningless patterns and run very slowly.

An efficient method to accelerate speed of parser is to compress
patterns set (combine similar patterns) and delete some unimportant
patterns from the dictionary. But which patterns should be deleted
or kept back is really a big problem, and it will be the main issue
of the next section.

\section{Accelerating Methods}
In parsing experiments on Penn Treebank, for example using chart
parsing, it is found that parsers can not parse sentences in durable
time, because too many POS, nonterminal and syntactic pattern types
have been generated. Obviously it is unacceptable for real-time
systems.

There are mainly two solutions to this problem: using Compressed POS
Set and Patterns Pruning. In this section, both the two algorithms
will be discussed, and experiments will be performed to prove the
algorithms.

\subsection{Compressed POS Set}
Compressed POS Set is a set of POS in which some POS in the full POS
set have been combined, in order to decrease the number of different
POS types. For example, we can combine ``NNS" and ``NNP", so
patterns ``NP $\rightarrow$ NNS" and ``NP $\rightarrow$ NNP" will be
combined into ``NP $\rightarrow$ NN". In syntactic patterns, both
terminal characters (e.g. NNS, VBD) and nonterminal characters (e.g.
NP, VP) are used, so we have to compress both of them to decrease
the amount of pattern types and then accelerate the parsing speed.

First, we combine terminals. For Penn Treebank style POS Set,
compressed POS Set in table \ref{Tab:Compressed} is applied to
decrease POS number. The POS in column ``Original" will be combined
into the POS in column ``Compressed". Thanks to the effect of this
table, the number of terminals has decreased from 46 to 27.

\begin{table}
  \centering
  \begin{tabular}{cc|cc}
  \hline
  Original & Compressed & Original & Compressed \\
  \hline
  NNS & NN & PRP\$& PRP\\
  NNP & NN & WP   & WDT\\
  NNPS& NN & WP\$ & WDT\\
  FW  & NN & WRB  & WDT\\
  VBD & VB & TO   & IN \\
  VBN & VB & JJR  & JJ \\
  VBG & VB & JJS  & JJ \\
  VBP & VB & RBR  & RB \\
  VBZ & VB & RBS  & RB \\
  \hline
\end{tabular}
  \caption{Compressed POS Set}\label{Tab:Compressed}
\end{table}

Second, we should combine nonterminals. For example, ``WHNP-22
$\rightarrow$ WDT", ``WHNP-23 $\rightarrow$ WDT", and ``WHNP-24
$\rightarrow$ WDT" are essentially the same, so they can be combined
into one pattern ``WHNP $\rightarrow$ WDT". So do patterns
``NP-SBJ-33 $\rightarrow$ DT NN", ``NP-SBJ-35 $\rightarrow$ DT NN"
and so on.

Table \ref{Tab:Comparison} shows the statistical data of differences
between using full POS set and compressed POS set. These data are
based on Penn Treebank 10\% version, in which there are total 10959
words and 94200 tag-tag pairs. In our experiments, it shows that the
amount of tag-tag pair types decreases to 341 using Compressed POS
Set, which is only 34\% of full POS set. The amount of pattern types
decreases to 4947, which is 61.8\% of full POS set. The amount of
nonterminal types decreases to 232, which is 35.2\% of full POS set.
The time elapsed in learning decreases by 16.7\%. Memory and disk
space occupied by learning result have also greatly decreased. All
these data prove that Compressed POS Set can effectively improve
tagging and parsing speed of a parser in a NLP module.

\begin{table}
  \centering
  \begin{tabular}{ccc}
    \hline
    Items & Full & Compressed\\
    \hline
    Terminal types & 46 & 27\\
    Dimension of HMM Array & 46 $\times$ 46 & 27$\times$27\\
    Tag-tag pair types  &  1003  &  341\\
    Word-tag pairs & 12726  & 11787\\
    Pattern types  & 8001   & 4947\\
    Nonterminal types & 659 & 232\\
    Time elapsed (ms) & 60515 & 50234\\
    \hline
  \end{tabular}
  \caption{Comparison of Different POS Sets}\label{Tab:Comparison}
\end{table}

\subsection{Syntactic Patterns Pruning}

Syntactic Patterns Pruning (SPP) is to delete some unimportant
patterns from pattern dictionary in order to save parsing time.

Compared with compressed POS Set, SPP is much more important for
accelerating parser speed. In parsing process, the seldom appearing
patterns waste much CPU time, but contribute nothing to improving
precision and recall. So in the case that is strict with processing
time and less important with precision, precision could decrease a
little by SPP in order to decrease the time elapsed in parsing. That
is a balance between precision and speed. Actually, in most cases,
users input short sentences instead of long sentences or complex
sentences in Penn Treebank, so the parsing precision will not
decrease greatly.

There are mainly three ways for SPP, which will be discussed as
follows.

\subsubsection{SPP on times thresholds}

\begin{figure}
  \includegraphics[bb=0 0 1201 900, scale=.2]{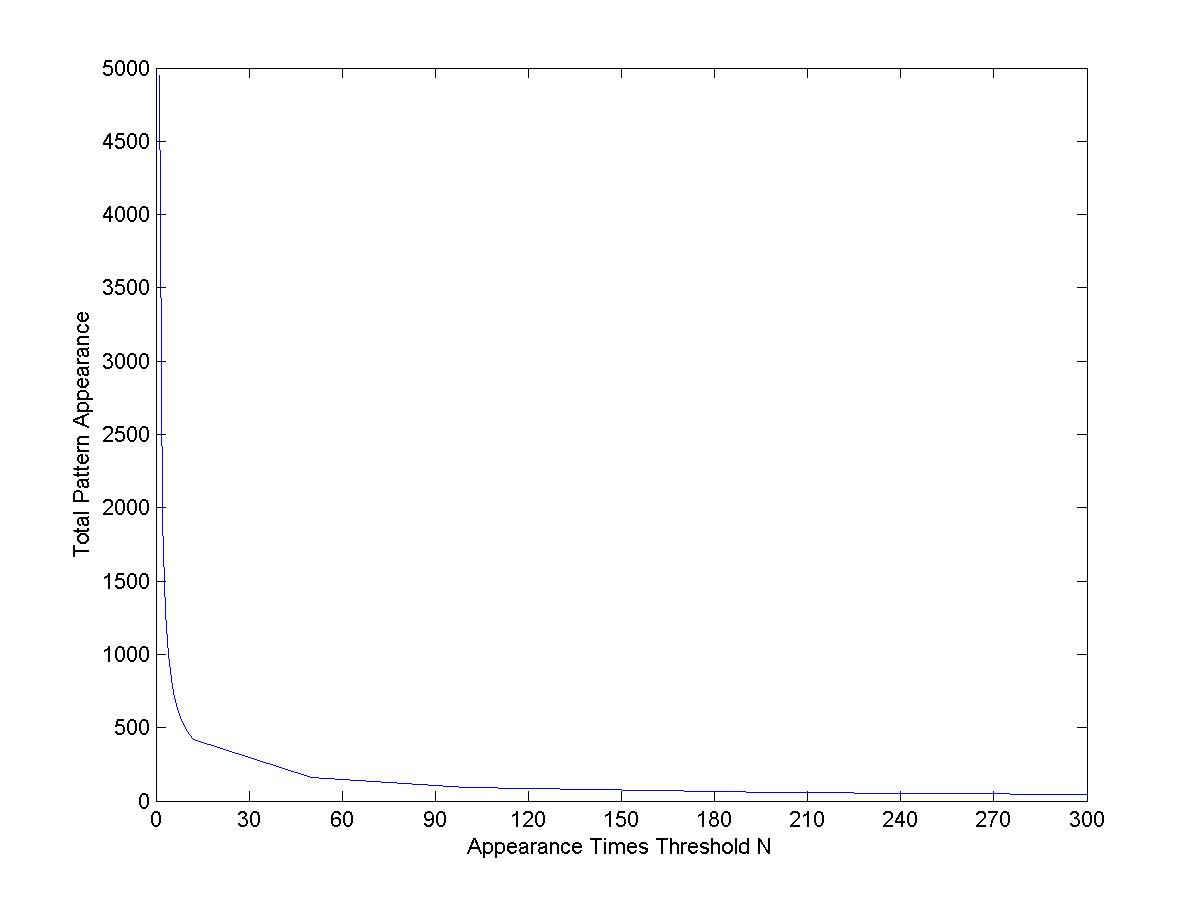}\\
  \caption{SPP on Appearance Times Threshold}\label{Fig:N}
\end{figure}

This method defines a threshold of pattern's appearance times, and
prunes patterns whose appearance times are less than the threshold.
Table \ref{Tab:SPP_Time} shows the relationship between the
threshold value and the amount of pattern types after pruning. And
the relationship is also depicted in Figure \ref{Fig:N} according to
the data in the table.

\begin{table}
  \centering
  \begin{tabular}{crrr}
    \hline
    Threshold N & PA & PT & NT \\
    \hline
    N=1   & 73460 & 4947 & 232 \\
    N=5   & 67623 & 826  & 88 \\
    N=8   & 66126 & 567  & 75 \\
    N=10  & 65315 & 472  & 69 \\
    N=12  & 64791 & 422  & 67 \\
    N=50  & 58675 & 160  & 37 \\
    N=60  & 57528 & 139  & 33 \\
    N=100 & 53704 & 89   & 28 \\
    N=300 & 45939 & 44   & 15\\
    \hline
  \end{tabular}\\

  PA: Appearance Times of remaining Patterns.\\
  PT: amount of remaining Patterns Types.\\
  NT: amount of Nonterminal Types.\\

  \caption{SPP on Appearance Times Threshold}\label{Tab:SPP_Time}
\end{table}

According to the data in the Table \ref{Tab:SPP_Time}, along with
the elevation of threshold, the amount of both pattern types and
nonterminal types decrease obviously, but the amount of patterns
appearance times decreases only a little. For example, at the point
$N=50$, the amount of pattern types decreases to 3.23\% of all
pattern types, and the amount of nonterminal types decreases to
15.9\% of all types, but the amount of pattern appearance times only
decreases to 80\% of all patterns.

The reason is that many patterns seldom appear, maybe only one or
two times, and these patterns can not greatly improve the precision
of the syntactic parser, but waste a lot of processing time. As a
result, these patterns should be removed from pattern dictionary in
order to improve speed.

In Figure \ref{Fig:N}, in N section $[15,50]$, the curve starts to
change much more smoothly. Our experiment in the later section shows
that the precision of the parser is still acceptable at the point
$N=50$.

After pruning, the remaining patterns are mainly like ``$NP
\rightarrow \gamma$" and ``$VP \rightarrow \gamma$", because NP and
VP appear in the corpus much more frequently than other nonterminals
do.

\subsubsection{SPP on probability threshold}
\begin{figure}[ht]
  \includegraphics[bb=0 0 1201 900, scale=.2]{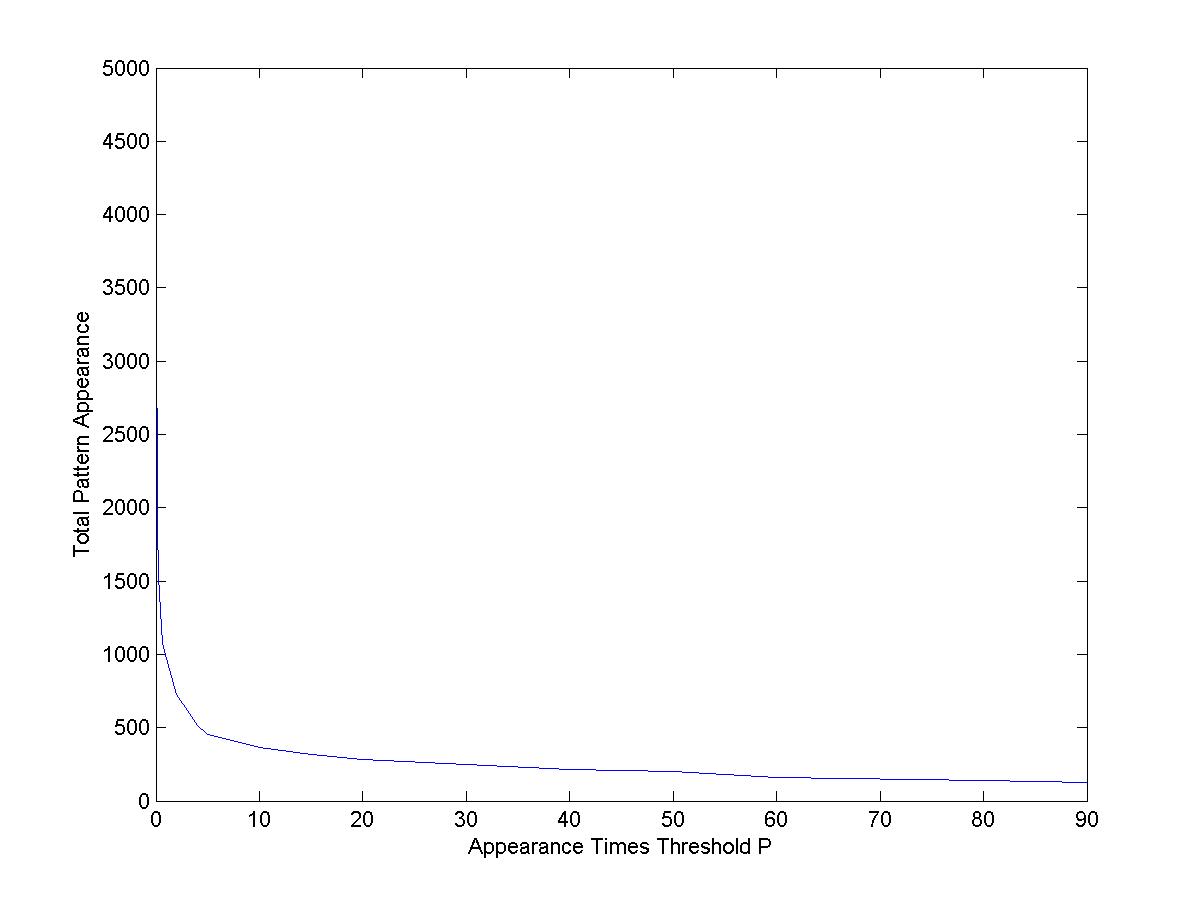}\\
  \caption{SPP on Probability Threshold}\label{Fig:P}
\end{figure}

This method defines a threshold of pattern appearance probability,
and prunes patterns whose appearance probability is smaller than the
threshold. Table \ref{Tab:SPP_Pro} shows the relationship between
the threshold and the amount of pattern types. And the relationship
is also shown in Figure \ref{Fig:P} according to the data in the
table.

\begin{table}
  \centering
  \begin{tabular}{crrr}
    \hline
    Threshold P & PA & PT & NT\\
    \hline
    P=0.00\%   & 73460 &  4947 & 232\\
    P=0.60\%   & 60459 &  1081 & 232\\
    P=2.00\%   & 51568 &  728  & 232\\
    P=5.00\%   & 45784 &  457  & 232\\
    P=10.00\%  & 35057 &  368  & 232\\
    P=15.00\%  & 26042 &  315  & 229\\
    P=20.00\%  & 23573 &  284  & 223\\
    P=40.00\%  & 17071 &  213  & 198\\
    P=80.00\%  & 10329 &  139  & 139\\
    \hline
  \end{tabular}
  \caption{SPP on Probability Threshold}\label{Tab:SPP_Pro}
\end{table}

According to the data in the Table \ref{Tab:SPP_Pro}, along with the
elevation of threshold, the amount of pattern types decreases
greatly, and the amount of patterns appearance times also decreases,
but the amount of nonterminal types hardly decreases, especially in
the section $[0,10\%]$ where it does not decrease at all. For
example, at the point $P=10\%$, the amount of pattern types
decreases to 7.43\% of all pattern types, and the amount of patterns
appearance times decreases to 47.7\% of all patterns, but the amount
of nonterminal types does not decrease.

The reason is that, the sum of the probabilities of all the patterns
with the same left side equals to 1:

\begin{equation}\label{Equ:Sum}
\sum_\zeta P(N^j \rightarrow \zeta) = 1
\end{equation}

As a result, when pruning on appearance probabilities threshold, if
the threshold P is small, nonterminals will not be pruned, such as
at the point $P=10\%$. But when the threshold is elevated, the
patterns with the same left side and different right side, in which
different constitutions of right side appear comparatively, will be
pruned first. Unfortunately, these patterns are always the most
common and important patterns. For example, an unimportant pattern
only appears once, then its probability is 100\%, and it will not be
pruned. So, nonterminals missing should be avoided if this method is
adopted. In other words, a low threshold should be defined.

In the Figure, in P section $[5,10]$, the curve starts to change
much more smoothly. Our experiment in later section shows that the
precision of syntactic parser is still acceptable at point $P=5$.

\subsubsection{Mixed SPP}
When we use SPP on appearance probability threshold, a great number
of syntactic patterns have to be reserved in order not to miss any
nonterminals, which will slow down speed of parser. So Mixed SPP
method, which prunes on both appearance times threshold and
probability threshold, could be adopted to keep the advantages of
both methods.

For example, if patterns which appear less than 30 times or have a
probability of less than 5\% are pruned, there are 44 nonterminals
and 44226 patterns left, which belong to 79 pattern types. In this
case, a syntactic parser can hardly correctly parse very long
sentences, but can effectively parse short sentences, which actually
are most frequently used by users, very fast and precisely. For
example, simple sentences could be parsed correctly, such as ``The
journal will report events of the past century" and ``I want to find
a job" etc.

Table \ref{Tab:SPP_Mix} shows the relationship between the values of
two thresholds and the amount of pattern types.

\begin{table}
  \centering
  \begin{tabular}{ccccc}
    \hline
    Threshold N & Threshold P & PA & PT & NT\\
    \hline
    30 & 5\% & 44226 &  79 & 44\\
    20 & 5\% & 44650 &  96 & 54\\
    10 & 5\% & 45081 & 125 & 69\\
    10 & 4\% & 46514 & 137 & 69\\
    10 & 3\% & 47049 & 147 & 69\\
    10 & 2\% & 50345 & 170 & 69\\
    \hline
  \end{tabular}
  \caption{Mixed SPP}\label{Tab:SPP_Mix}
\end{table}

\section{Evaluation}
In previous sections, it has been demonstrated that Compressed POS
Set and SPP can accelerate parsers' speed very effectively. But
because of the different definitions of thresholds N and P, parsers
with different thresholds will perform differently. A certain
evaluation method should be proposed to evaluate different $<N,P>$
pairs for Mixed SPP Method.

In this section, two new evaluation factors are defined to describe
parsers' efficiency. The higher its value is, the faster and more
accurate the parser is.

\subsection{PT Factor}
As what has been discussed in the previous sections, in order to
accelerate parser, we should keep a balance between speed and
precision, through using Compressed POS Set and SPP algorithm.

To get a high score in efficiency evaluation, parsers should process
sentences correctly as many as possible in a time unit. Parameter
$\mu$ is defined to represent this concept:

\begin{equation}
\mu = \frac{C^+}{T}
\end{equation}

In the formula, $C^+$ represents the amount of syntactic patterns
correctly parsed by parser, and $T$ represents the time elapsed in
parsing.

Assume that parameter $C$ represents the total amount of parsed
syntactic patterns, including both correctly and incorrectly parsed
patterns. Because tests on different parsers are based on the same
test set, $C$ values are equal. As a result, the ratio of
$\mu_1,\mu_2$ equals to the following:

\begin{equation}\label{Equ:muDiv}
\frac{\mu_1}{\mu_2} = \frac{C^+_1}{T_1} \frac{T_2}{C^+_2}\\
= \frac{C^+_1}{T_1 C} \frac{T_2 C}{C^+_2}
\end{equation}

Besides, the precision of a system, defined as $P$, should be
computed as the following:

\begin{equation}
P = \frac{C^+}{C}
\end{equation}

So, formula \ref{Equ:muDiv} is transformed into the following
formula:

\begin{equation}
\frac{\mu_1}{\mu_2} = \frac{P_1}{T_1} / \frac{P_2}{T_2}
\end{equation}

Here, magnitude of $\mu$ is decided by the ratio of precision and
time. So, factor $PT$ is proposed to evaluate time efficiency of a
parser, which is defined as the following:

\begin{equation}
PT = \frac{P}{T}
\end{equation}

where $P$ represents precision of parsing, and $T$ represents time
elapsed in parsing. An efficient parser should be of higher $PT$
value.

\subsection{RT Factor}
Obviously, $PT$ factor can evaluate parsers effectively. However,
this situation exists: as a result of too much pruning, along with
great decrease of time elapsed, precision also greatly decreases. In
this situation, $PT$ value is nearly the same as parser with high
precision. To solve this problem, the balance between recall and
speed should also be evaluated. That means, parser should correctly
recall as many syntactic patterns as possible in a time unit.
Variable $\lambda$ is defined to represent this concept:

\begin{equation}
\lambda = \frac{C^+_r}{T}
\end{equation}

where, $C^+_r$ represents the amount of syntactic patterns correctly
recalled by parser, and $T$ represents time elapsed in parsing.
Also, tests on different parsers are based on the same test set, so
$C$ values are equal. As a result, the ratio of
$\lambda_1,\lambda_2$ equals to the following:

\begin{equation}\label{Equ:lambdaDiv}
\frac{\lambda_1}{\lambda_2} = \frac{C^+_{r1}}{T_1}
\frac{T_2}{C^+_{r2}} = \frac{C^+_{r1}}{T_1 C} \frac{T_2 C}{C^+_{r2}}
\end{equation}

The recall rate of the system, which is defined as $R$, should be
computed as the following:

\begin{equation}
R = \frac{C^+_r}{C}
\end{equation}

So, formula \ref{Equ:lambdaDiv} is transformed into the following
formula:

\begin{equation}
\frac{\lambda_1}{\lambda_2} = \frac{R_1}{T_1} / \frac{R_2}{T_2}
\end{equation}

Here, magnitude of $\lambda$ is decided by the ratio of recall rate
and time elapsed. So, factor $RT$ is also proposed to evaluating
time efficiency of parser, which is defined as the following:

\begin{equation}
RT = \frac{R}{T}
\end{equation}

In this formula, $R$ represents recall rate of parser, and $T$
represents time elapsed in parsing. $RT$ represents the amount of
correctly recalled patterns in a time unit. An efficient parser
should be of higher $RT$ value also, not only higher $PT$ value.

The following example shows how to compute $PT$ and $RT$ values. A
test, in which compressed POS set is used and the patterns that
appears less than 50 times has been pruned, has been performed on
the first 5854 lines of Penn Treebank. The precision of syntactic
parsing is: $247/691=35.7\%$, and the recall rate is:
$247/761=32.5\%$, 4048 seconds elapsed in the test. So $PT$ and $RT$
values could be calculated as follows:

\begin{equation}
PT=35.7/4048=0.0088
\end{equation}
\begin{equation}
RT=32.5/4048=0.0080
\end{equation}

Our experiment in the next section will discuss how to choose
parameters in order to increase $PT$ and $RT$ values.

\section{Experiments}
Our experiments are performed on 10\% version of Penn Treebank,
obtained from NLTK \cite{Loper2004}. Compressed POS Set is adopted
in all the following experiments.

Firstly, precision of our tagger is tested on the first 9988 lines
of Penn Treebank, and the result is 9479/9784 = 96.88\%.

Then, parsers with different thresholds are tested on the first 5854
lines:

(1) Threshold N=50, Precision = 247/691 = 35.7\%, Recall = 247/761 =
32.5\%.

(2) Threshold N=60, Precision = 235/662 = 35.5\%, Recall = 235/761 =
30.9\%.

(3) Threshold P=60\%, Precision = 0, Recall = 0.

(4) Threshold P=5\%, Precision = 30/94 = 31.9\%, Recall = 30/761 =
3.9\%.

(5) Threshold N=10, P=2\%, Precision = 72/208 = 34.6\%, Recall =
72/761 = 9.5\%. Although test (5) gets a lower recall than (1), its
parsing speed is far faster than (1).

\begin{table*}
  \centering
  \begin{tabular}{ccccr|c}
    \hline
    Threshold N & Threshold P & Pattern Types & Precision  & Time Elapsed & PT Value\\
    \hline
    50 & 0\% & 160 & 35.7\% & 4048.578s & 0.0088\\
    60 & 0\% & 139 & 35.5\% & 3091.563s & 0.0115\\
    0  & 60\%& 163 & 0      &  250.421s & 0\\
    0  & 5\% & 457 & 31.9\% &  249.359s & 0.1281\\
    10 & 2\% & 170 & 34.6\% &  278.329s & 0.1245\\
    \hline
  \end{tabular}
  \caption{PT Value after Pruning}\label{Tab:PT}
\end{table*}

\begin{table*}
  \centering
  \begin{tabular}{ccccr|c}
    \hline
    Threshold N & Threshold P & Pattern Types & Recall  & Time Elapsed & RT Value\\
    \hline
    50 & 0\%  & 160 & 32.5\% & 4048.578 & 0.0080\\
    60 & 0\%  & 139 & 30.9\% & 3091.563 & 0.0100\\
    0  & 60\% & 163 & 0      & 250.421  & 0\\
    0  & 5\%  & 457 & 3.9\%  & 249.359  & 0.0157\\
    10 & 2\%  & 170 & 9.5\%  & 278.329  & 0.0342\\
    \hline
  \end{tabular}
  \caption{RT Value after Pruning}\label{Tab:RT}
\end{table*}

All the related data of the tests is summarized in Table
\ref{Tab:PT} and Table \ref{Tab:RT}. Of the five tests, test (5), in
which the amount of pattern types is about the same as other tests
or even less than the others, has nearly both the highest $PT$ and
$RT$ values. So thresholds in test (5) are the best parameters in
the test for system which is strict with processing time.

But in the cases that systems are not strict with processing time,
the pruning algorithm with lower $PT$ and $RT$ value but higher
precision and recall should be adopted. For example, in the case of
ignoring small differences of precision and recall between test (1)
and test (2), test (2) is better than test(1).

Actually, in real systems, users always input short sentences, so
precision and speed will be much higher than these experiments.


\section{Future Work}
In the future, more accelerating algorithms should be proposed to
improve parser, which will greatly promote the application of NLP
technology, especially in real-time systems. These methods may
include:

(1) Determine the importance level of syntactic patterns by their
content and constitution. For example some patterns frequently
appear in written English corpus but seldom appear in oral English
or user input, so these patterns should be removed from the
dictionary and that will not influence precision and recall.

(2) Instead of chart parsing, a new parsing algorithm may be
proposed for the new speed demands. In the new algorithm, the
disadvantages brought by step ``predict" should be avoided.

As to the evaluation, actually, in different application background,
different evaluation factors should be defined for the specified
circumstance. In other words, Precision and Recall are not the only
things we should pay attention to.

\subsection*{Acknowledgments}
Here we thank Prof. Yuxi Fu, who supports our project and gives us
lots of valuable advice.

\end{document}